%% file: main.tex
\documentclass[letterpaper, 10 pt, conference]{ieeeconf}  

\IEEEoverridecommandlockouts                              

\usepackage{times}
\usepackage{epsfig}
\usepackage{graphicx}
\usepackage{amsmath}
\usepackage{amssymb}
\usepackage{algorithm}
\usepackage[noend]{algpseudocode}

\usepackage{subcaption}
\DeclareCaptionSubType*[Alph]{table}
\DeclareCaptionLabelFormat{mystyle}{Table~\bothIfFirst{#1}{ ̃}#2}
\usepackage{rotating}

\usepackage{color}
\usepackage{xcolor}
\usepackage{comment}

\title{\LARGE \bf
Learning Novel Objects Continually Through Curiosity
}

\author{Ali Ayub$^{1}$ and Alan R. Wagner$^{2}$
\thanks{$^{1}$Department of Electrical Engineering,
        The Pennsylvania State University, State College, PA 16802, USA
        {\tt\small aja5755@psu.edu}}%
\thanks{$^{2}$Department of Aerospace Engineering, The Pennsylvania State University,
        State College, PA 16802, USA
        {\tt\small alan.r.wagner@psu.edu}}
}

\begin{document}

\maketitle
\thispagestyle{empty}
\pagestyle{empty}

\input{abstract.tex}
\input{introduction.tex}

\input{methodology.tex}

\input{experiments.tex}
\input{conclusion.tex}
\section*{Acknowledgment}
\noindent This work was partially supported by the Air Force Office of Sponsored Research contract FA9550-17-1-0017.


{\small
\bibliographystyle{IEEEtran.bst}
\bibliography{main}
}

\end{document}

%% file: abstract.tex
\begin{abstract}
\label{sec:Abstract}
Children learn continually by asking questions about the concepts they are most curious about. With robots becoming an integral part of our society, they must also learn unknown concepts continually by asking humans questions. The paper analyzes a recent state-of-the-art approach for continual learning. The paper further develops a self-supervised technique to find most of the uncertain objects in an environment by utilizing the cluster representation of the previously learned classes. We test our approach on a benchmark dataset for continual learning on robots. Our results show that our curiosity-driven continual learning approach beats random sampling and softmax-based uncertainty sampling in terms of classification accuracy and the total number of classes learned. 
   
\end{abstract}

%% file: introduction.tex
\section{Introduction}
\label{sec:introduction}







\noindent Learning in children is driven by curiosity: children actively ask teachers questions about the concepts that they are unfamiliar with \cite{piaget77,richards03}. This learning process is extremely efficient, the child only asks about the information he/she is missing and the teacher does not have to answer unnecessary questions \cite{michelle07}. Using curiosity, children (and all humans) continue to expand their knowledge over time by learning new concepts without the need to relearn most previous concepts.

With robots increasingly becoming an integral part of the society for a  variety of different roles, such as household robots \cite{Matari17}, they should also learn by asking questions from humans. Since, it is difficult for human teachers to answer enormous numbers of questions, robots must learn in a curiosity driven manner, i.e. they should prioritize and only ask about the concepts they are most uncertain about by asking simple questions from humans. Further, since robots have limited on-board memory it is impossible for them to store raw data belonging to all of the previously learned concepts. Hence, robots must use curiosity to learn continually over time, without the need to relearn previously learned concepts. In this paper, we consider a method that could possibly be used to allow robots to explore and learn novel, previously unseen objects.  

Deep learning has achieved remarkable success in object recognition~\cite{He_2016_CVPR}. Yet, a deep neural network requires a large  corpus of high-quality labeled data in a single batch \cite{french19}, which makes it impractical for real-world robotics applications. In recent years, attempts have been made to develop active learning techniques to reduce the cost of labelling \cite{Beluch_2018_CVPR,Gal17,Siddiqui_2020_CVPR,Yoo_2019_CVPR,Shen_2019_ICCV}. All of these techniques use uncertainty sampling to request labels for the most uncertain objects. Most of these techniques, however, train a base neural network model with a specific loss to measure the uncertainty of samples using the training set. When learning from new unlabeled data, the model first determines which samples should be labelled and then retrains the network on a combination of previously learned and newly labeled data. Hence, these techniques cannot learn in a continual manner without requiring any old data and will suffer from \textit{catastrophic forgetting}: a phenomenon in which the model forgets the previously learned classes when learning new classes and the overall classification accuracy decreases drastically. 

To overcome catastrophic forgetting, many different deep learning techniques have been proposed for continual learning \cite{Ayub_iclr20,Rebuffi_2017_CVPR,Wu_2019_CVPR,Ayub_2020_CVPR_Workshops,Castro_2018_ECCV,Dhar_2019_CVPR}. Most continual learning techniques, however, require the training data of the earlier learned object classes when learning new classes \cite{Rebuffi_2017_CVPR,Wu_2019_CVPR,Castro_2018_ECCV}. Further, most continual learning techniques require the complete labeled data belonging to a class in a single increment which makes them impractical for robotics applications where data belonging to earlier learned classes and new classes is available in a streaming manner. Lastly, all of the continual learning techniques assume that all the data is labeled in each increment, hence they cannot learn when most of the data is unlabelled and they have to prioritize which data samples to learn about. 

\begin{figure*}
\centering
\includegraphics[scale=0.5]{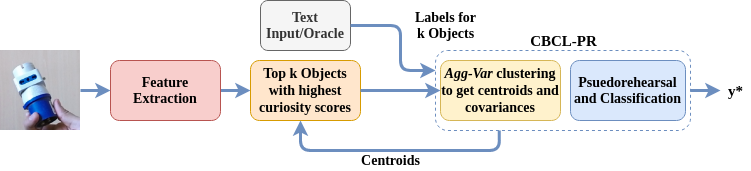}
\caption{\small Overall architecture of our approach. The model gets new unlabeled objects (example image on the left) and extracts CNN features for all the images of the objects. Previously learned clusters are used to calculate curiosity scores for all the objects and top $k$ objects are labeled by the oracle. \textit{Agg-Var} clustering is then called on the labeled object features to get updated centroids and covariance matrices which are then used during pseudo-rehearsal and shallow classifier training.}
\label{fig:framework}
\end{figure*}

In this paper, we propose an approach that can be used for continually learning novel objects through curiosity. Unlike previous approaches our approach continually learns through curiosity without suffering from catastrophic forgetting. We analyze our previously proposed approach for continual learning termed Centroid-Based Concept Learning with Pseudorehearsal (CBCL-PR) \cite{Ayub_2020_CVPR_Workshops,Ayub_cbclpr_IROS_21,Ayub_ICSR20} and show that the cluster representations learned from the previous data can be used to learn unseen objects through curiosity. We perform a preliminary experiment on a benchmark dataset for continually learning objects through curiosity. Results on the dataset show that our approach is more efficient than simple random sampling and softmax-based uncertainty sampling for learning object classes continually.

%% file: methodology.tex
\section{Curiosity-Driven Online Learning}
\label{sec:methodology}

\noindent Figure \ref{fig:framework} shows the overall architecture of our approach. The system can learn object classes continually using CBCL-PR \cite{Ayub_cbclpr_IROS_21}. The clusters generated by CBCL-PR can be used for assigning uncertainty scores to new unlabeled objects. The subsections below describes all the components of our approach in detail.

\subsection{Centroid-Based Concept Learning and Pseudorehearsal}
\label{sec:cbcl_pr}

CBCL-PR \cite{Ayub_cbclpr_IROS_21} is composed of two modules: 1) \textit{Agg-Var} clustering, and 2) a pseudo-rehearsal and classification module. In the learning phase, once the human provides the robot with the training examples for a new class, the first step in CBCL-PR is the generation of feature vectors from the images of the new class using a fixed feature extractor. For object classification, we use ResNet-18 \cite{He_2016_CVPR} pre-trained on the ImageNet dataset~\cite{Krizhevsky12} as the feature extractor. 

\subsubsection{\textit{Agg-Var} Clustering}
\label{sec:agg_var}
In the case of continual learning, for each new image class $y$, CBCL-PR clusters all of the training images in the class. \textit{Agg-Var} clustering \cite{Ayub_2020_CVPR_Workshops}, a cognitively-inspired approach, begins by creating one centroid from the first image in the training set of class $y$. Next, for each image in the training set of the class, feature vector $x_i^y$ (for the $i$-th image) is generated and compared using the Euclidean distance to all of the centroids for the class $y$. If the distance of $x_i^y$ to the closest centroid is below a pre-defined distance threshold $D$, the closest centroid is updated by calculating a weighted mean of the centroid and the feature vector $x_i^y$. If the distance between the $i$th image and the closest centroid is greater than the distance threshold $D$, a new centroid is created for class $y$ and equated to the feature vector $x_i^y$ of the $i$-th image. 

CBCL-PR also calculates the covariance matrices related to each centroid using the feature vectors of all the images clustered in the centroid. The result of this process is a collection containing a set of centroids, $C^y = \{c_1^y, ..., c_{N^*_y}^y\}$, and covariance matrices, $\sum^y = \{\sigma_1^y,...,\sigma_{N^*_y}^y\}$, for the class $y$ where $N^*_y$ is the number of centroids for class $y$. This process is applied to the sample set $X^y$ of each class continually once they become available to get a collection of centroids $C  = C^1, C^2, ..., C^N$ and covariance matrices $\sum = \sum^1, \sum^2, ..., \sum^N$ for all $N$ classes in a dataset. 

\subsubsection{Pseudorehearsal}
\label{sec:pseudorehearsal}
For classification of images from all the classes seen so far, CBCL-PR trains a shallow neural network composed of a linear layer trained with a softmax loss on examples from the current increment and pseudo-exemplars of the old classes (pseudorehearsal). Pseudo-exemplars are not real exemplars of a class, rather they are generated based on the class statistics to best resemble the actual exemplars. 

CBCL-PR uses each of the centroids $c_i^y$ and covariance matrices $\sigma_i^y$ of class $y$ as parameters of a Gaussian distribution. 
The Gaussian distribution is sampled to generate the same number of pseudo-exemplars as the total number of exemplars represented by the centroid/covariance matrix pair $c_i^y$,$\sigma_i^y$. This process leads to a total of $N_y$ number of pseudo-exemplars for class $y$. In this way, the network is trained on $N_y$ number of examples for each class $y$ in each increment. For classification of a test image, the feature vector $x$ of the image is first generated using the feature extractor. Next, this feature vector is passed through the shallow network to produce a predicted label $y^*$.

\begin{figure}[t]
\centering
\includegraphics[width=1.0\linewidth]{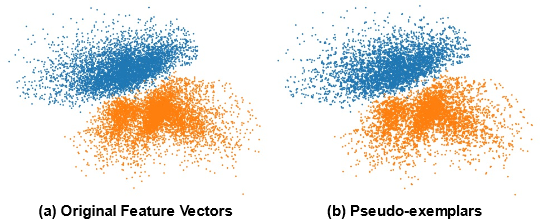}
\caption{\small Comparison of the original feature space (11379 feature vectors) with the pseudo-exemplars generated by 1000 clusters for 2 classes in the MNIST dataset.}
\label{fig:mnist_example}
\end{figure}

\subsection{An Example on the MNIST Dataset}
To show insight into CBCL-PR and its use for curiosity-driven learning, we performed a small experiment on the MNIST dataset \cite{Lechun98}. We applied CBCL-PR on two classes from the MNIST dataset with 11379 images. We transformed the embedding (feature vector) size from 512 to 2 to visualize the feature vectors it in the 2D space. CBCL-PR generated 1000 clusters (centroids and covariance matrices) for the two classes. The memory required by the clusters is $\sim$17\% of the memory required by the original feature vectors. Figure \ref{fig:mnist_example} shows the original feature vectors and the pseudo-exemplars generated by CBCL-PR for the two classes in the MNIST dataset. The 2D space covered by the pseudo-exemplars looks almost the same as the original feature vectors, which shows that CBCL-PR can learn the complex distributions of the image data using clusters (Gaussian distributions). We believe that the information contained in the clusters of the old classes can be used to determine if the new, previously unseen samples (objects) belong to any of the clusters.




\subsection{Self-Supervised Curiosity Scores}
\label{sec:curiosity_module}
The curiosity score attached to each new object is dependent on how much it is unknown to the robot. For example, if a new object belongs to a completely new class the robot should be more curious about learning about that object rather than another object that belongs to a class already learned by the robot. To find the curiosity scores attached to each new object, we present a self-supervised technique that uses the clusters of the classes learned through CBCL-PR.

During an increment, the robot gets $m$ new objects and it captures $n_j$ images for each object $j$ to capture different views of the object. To learn the labels/ground truth of the objects, the robot can ask a human partner for assistance. However, it cannot ask about all $m$ objects because of people having limited attention span. Humans are typically unwilling to answer a large number of questions at one time. Assume that the robot can only ask $k<m$ questions, hence it has to prioritize which of the $k$ objects it should ask about. Before finding the curiosity scores, we first find the feature vectors for all the images of $m$ objects by passing them through the CNN feature extractor used for CBCL-PR. To prioritize objects, we find the curiosity score ($\mathcal{A}_j$) of each object $j$ as: 

\begin{equation}
    \mathcal{A}_j = \frac{m}{\sum_{i=1}^m {p_j^i}}
\end{equation}

\noindent where, $p_j^i$ is the probability of $i$th image of the $j$th object to belong to the most probable cluster. The term $\mathcal{A}_j$ is the inverse of the average probabilities of all the images of object $j$ to belong to the most probable clusters. A higher value of $\mathcal{A}_j$ means higher uncertainty, hence higher curiosity score, for object $j$. Using the curiosity scores, the top $k$ objects with the higher curiosity scores are chosen to be labeled.


%% file: experiments.tex
\section{Experiments}
\label{sec:experiments}
\noindent We perform a preliminary experiment on the CORe-50 dataset \cite{lomonaco17} to evaluate our approach for continually learning novel objects through curiosity. First, we present the CORe-50 dataset and the implementation details and then compare our approach against random sampling and softmax based uncertainty sampling.

\begin{figure*}[t]
\centering
\includegraphics[scale=0.35]{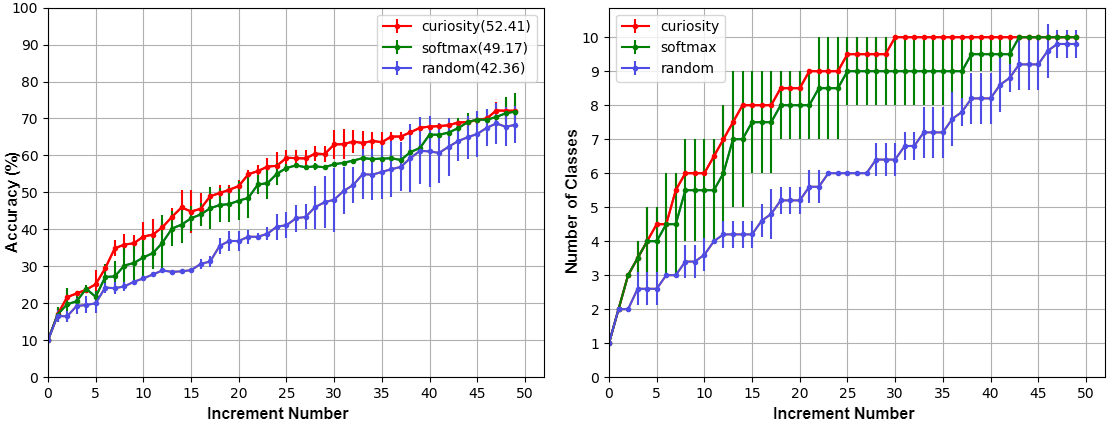}
\caption{\small Comparison of our method (red curve) to the softmax based sampling (green curve) and random sampling (blue curve) in terms of classification accuracy (left) and number of classes learned (right) in 50 increments for 10 object level classes on the CORe-50 dataset. Average incremental accuracy is reported in parenthesis. The curves show average and standard deviation of 5 experiments with random seeds. (Best viewed in color)}
\label{fig:curiosity_experiment}
\end{figure*}

\subsection{CORe-50 Dataset}
CORe-50 is an object recognition dataset specifically designed for continual learning that contains household objects from ten different object categories. CORe50 is realistic in that each object class is made of 5 object instances. Further, each object class is captured with 11 different background environments (indoor, outdoor etc.) from 11 different recording sessions. Each session is composed of an approximately 15 second video clip recorded at 20 fps. We used the cropped 128$\times$128 images and used sessions 3, 7 and 10 as the test set provided by the authors in \cite{lomonaco17}. It should be noted that the test set contains all the samples of all the classes, however the model learns object classes continually. Hence, in the initial increments the performance of any model is extremely poor as shown in \cite{lomonaco17}. We performed experiments with 10 object level classes.

\subsection{Implementation Details}
We used the Pytorch deep learning framework \cite{torch19} for implementation and training of all neural network models. We used ResNet-18 \cite{He_2016_CVPR} pre-trained on ImageNet \cite{Krizhevsky12} as a feature extractor for CBCL-PR. Also, we only used the diagonal entries of the covariance matrix to keep the memory budget from growing drastically.

We combined all 8 training sessions in the CORe-50 dataset to generate 400 training object instances. In each increment, we sampled 5 object instances and allowed the model to learn the label of 1 out of 5 object instances. Hence, we allowed the model to learn all objects in 400 increments with one object learned in each increment. To the best of our knowledge, there are no other continual learning approaches that are specifically designed for curiosity-driven learning. Hence, to evaluate our self-supervised curiosity-driven approach we compared against random sampling and softmax based uncertainty sampling with CBCL-PR as the online learning algorithm. For random sampling, the system randomly selects one object instance out of 5 instances in an increment and learns it using CBCL-PR. Random sampling is similar to traditional an online learning setting when the robot learns from new data randomly without deciding what it should learn next. For the softmax based curiosity scores, we found the softmax output of each image in an object instance using CBCL-PR. Then, we took the average of the maximum probability in the softmax output of each image of the object instance. The object instance with the highest average probability score was chosen to be labeled. We used the test set accuracy in each increment as the evaluation metric. We only report the accuracies for 50 increments for this experiment because all three approaches learn all 10 classes by 50 increments and the advantage of curiosity-driven learning decreases for all the approaches leading to similar accuracy after approximately 50 increments since all of them use CBCL-PR as the online learning algorithm. We also report average incremental accuracy. As another metric to evaluate curiosity-driven online learning, we report how many increments the three approaches require to learn all 10 classes. 

For robustness, the experiment were performed 5 times with random seeds. We report the average and standard deviation of the accuracies. Hyperparameters $D$ is set to 17.5 for all increments.

\subsection{Curiosity-Driven Continual Learning Evaluation}
Figure \ref{fig:curiosity_experiment} (left) compares our curiosity-driven approach against random sampling and the softmax-based approach in terms of classification accuracy on the fixed test set for 10 classes. Our proposed approach beats both the softmax based and random sampling approach for all 50 increments. However, the difference in classification accuracy starts to decrease in later increments since all of the models learn all the classes as the number of increments increases. These results depict the ability of our approach to choose the most informative samples in each increment which makes it perform better than the other approaches, especially in the earlier increments. 

Figure \ref{fig:curiosity_experiment} (right) shows the total number of classes learned at each increment by the three approaches. Because of its ability to find the most uncertain samples, our proposed approach learns the maximum number of classes (10 classes) by the 30th increment. Softmax-based approach learns all the classes by 43rd increment while random sampling learns all the classes in 50 increments. These results show that our proposed approach allows the model to learn classes faster than other approaches. It should be noted that our curiosity-driven approach can only be applied with CBCL-PR as the continual learning algorithm. Other continual learning approaches like BiC \cite{Wu_2019_CVPR} have to use softmax-based uncertainty sampling, which will likely lead to inferior performance compared to our approach. 

%% file: conclusion.tex
\section{Conclusion}
\label{sec:conclusion}
\noindent This paper presents and evaluates a method for a novel problem: curiosity-driven continual learning of objects. Our paper demonstrates that curiosity-driven learning is highly efficient and helps the model learn the most uncertain object classes in earlier increments. We show that the cluster representations generated by our continual learning approach can be used to find curiosity scores for unseen objects. Finally, our experimental results demonstrate that  our approach leads to better performance than random sampling and softmax-based uncertainty sampling for online learning of objects. In our future work, we hope to develop better metrics for assigning curiosity-scores to novel objects using the clusters of the previous classes. We hope to apply our approach on a real robot wandering in an unconstrained environment for learning objects by asking questions from humans. 